\documentclass{article}
\usepackage[preprint]{corl_2026}
\usepackage[T1]{fontenc}
\usepackage[utf8]{inputenc}
\usepackage{amsmath,amssymb,booktabs,tabularx,array}
\usepackage{microtype,graphicx,wrapfig}
\hypersetup{hidelinks,pdfauthor={Jingxu Gu and Dongze Lian},pdftitle={Improving Fine-Grained Manipulation through Demonstration Expansion and Execution Design}}
\newcolumntype{Y}{>{\raggedright\arraybackslash}X}
\title{Improving Fine-Grained Manipulation through Demonstration Expansion and Execution Design}

\author{%
  Jingxu Gu \qquad Dongze Lian\thanks{Corresponding author.} \\
  {\normalfont Nanjing University} \\
  {\normalfont\texttt{\href{mailto:jingxugu@smail.nju.edu.cn}{jingxugu@smail.nju.edu.cn}}\quad
  \texttt{\href{mailto:liandz@nju.edu.cn}{liandz@nju.edu.cn}}}
}

\begin{document}
\setcounter{footnote}{1}
\maketitle
\setcounter{footnote}{0}
\begin{abstract}
Fine-grained manipulation requires a robot policy to follow instructions, remain robust to visual variations, and translate predicted actions into precise robot control. We present Demonstration Expansion and Execution Design (DEED), a framework for adapting a pretrained vision-language-action policy to the five MetaFine tasks through structured demonstration expansion and execution design. DEED generates demonstrations for different instructions from the same initial scene and renders each physical trajectory under different camera positions and lighting conditions while preserving its original states and action targets. A hierarchical sampling strategy balances instructions, trajectories, and visual conditions during supervised adaptation. At deployment, DEED converts predicted action chunks into robot commands using inverse kinematics, gripper mappings selected for each task, and closed-loop replanning. DEED achieves an official Overall score of 85.55\%. Fixed-policy evaluations show that performance varies with execution horizon, denoising budget, and gripper mapping.
\end{abstract}

\section{Introduction}

Language-conditioned manipulation requires a robot to identify the object, part, or motion specified by an instruction and translate it into precise spatial control~\citep{shridhar2022cliport,mees2021calvin}. Recent Vision-Language-Action (VLA) policies combine pretrained visual-language representations with robot action learning and have demonstrated broad transfer across tasks and robot embodiments~\citep{brohan2023rt2,kim2024openvla,black2024pi0,octo2024,openx2023}. However, fine-grained manipulation remains challenging because it jointly requires instruction-sensitive target selection, robustness to visual variations, and accurate conversion from predicted actions to robot control.

MetaFine~\citep{xu2026beyond,benchmark} evaluates these capabilities through five tasks covering part-level grasping, directional motion, switch toggling, object sorting, and letter insertion. Its evaluation considers three dimensions: \emph{perception} under camera and illumination changes, \emph{understanding} of instruction-conditioned behavior, and \emph{controlled behavior} during task execution.

Directly fine-tuning a pretrained VLA policy on the provided demonstrations leaves three practical gaps. First, observing different instructions only in different scenes can lead to scene-instruction confounding. Second, limited camera and illumination conditions provide insufficient visual coverage. Third, predicted action chunks may fail when the execution horizon, inverse kinematics conversion, or gripper semantics do not match the simulator interface. Addressing these gaps requires adapting both the demonstration distribution and the execution pipeline.

\begin{figure}[t]
\centering
\includegraphics[width=0.96\linewidth]{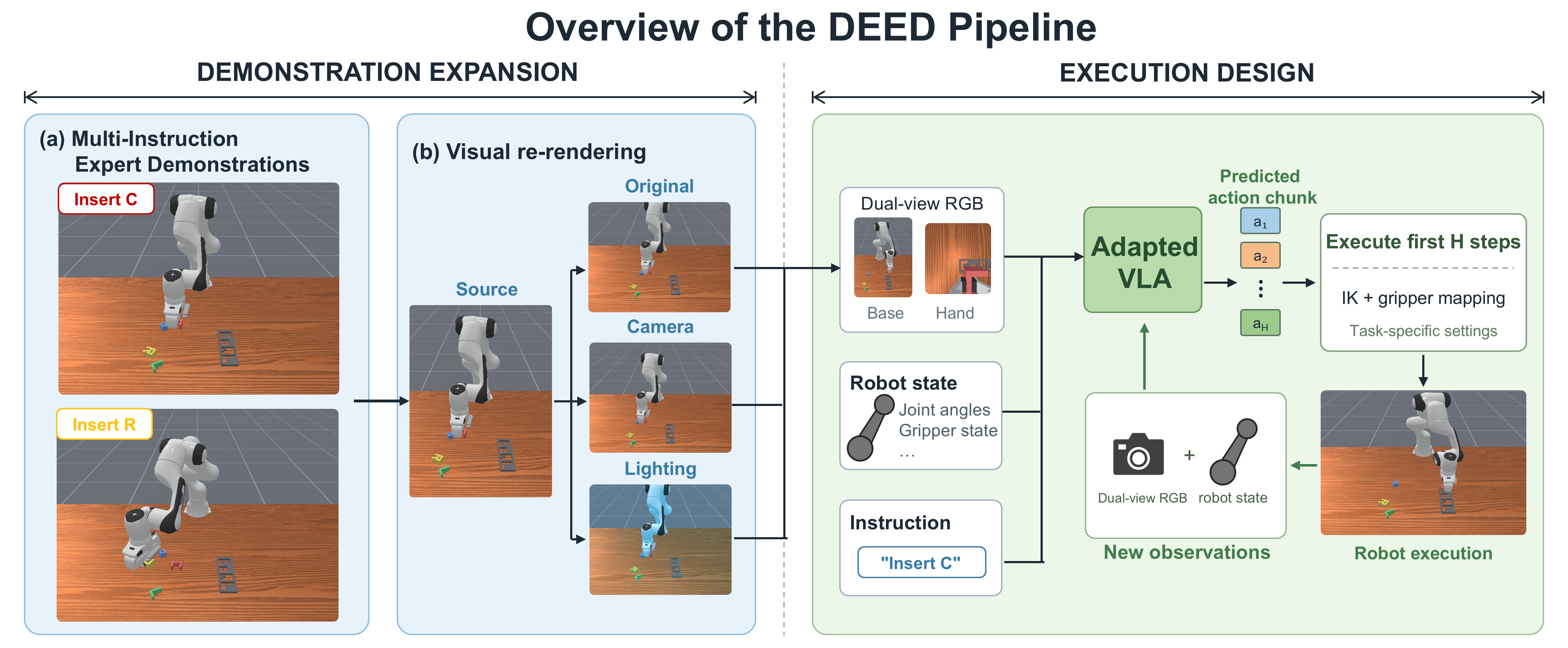}
\caption{Overview of DEED (T5 illustrated). Paired demonstrations and visual re-rendering support adaptation; deployment executes an $H$-step prefix of each $K$-step prediction before re-observation.}
\label{fig:pipeline}
\end{figure}

To address these gaps, DEED expands demonstrations across instructions and visual conditions, balances training exposure through hierarchical sampling, and adapts closed-loop execution to each task (Figure~\ref{fig:pipeline}).

Our contributions are: (1) A demonstration expansion strategy that varies instructions within the same scene and visual conditions for the same physical trajectory, supported by hierarchical sampling. (2) A task-aware execution pipeline that converts predicted action chunks into robot control through inverse kinematics, task-specific gripper mappings, and receding-horizon replanning. (3) A five-task evaluation achieving an official Overall score of 85.55\%, together with fixed-policy studies of key execution choices.

\section{Method}

\subsection{Instruction-paired demonstration expansion}
Automated demonstration generation can expand imitation-learning datasets~\citep{mandlekar2023mimicgen,nasiriany2024robocasa}.
Our inspection of the released T5 metadata found disjoint scene seeds across letter subsets~\citep{dataset}; pairing addresses the resulting opportunity for scene-based shortcuts.
Let $z$ index an initial scene with physical state $x_0^z$, and $\mathcal L_z$ denote its instruction family.
The official expert planner $\mathcal E$ constructs
\begin{equation}
 \tau_{z,\ell}=\mathcal E(x_0^z,\ell),\qquad \ell\in\mathcal L_z .
 \label{eq:pair}
\end{equation}
A saved physical trajectory is $\tau_{z,\ell}=(\ell,\{(x_t,s_t,a_t)\}_{t=0}^{T-1})$, where $T$ is its length, $x_t$ is the simulator state needed to reconstruct the scene (including robot and object configurations), $s_t$ is the policy's proprioceptive input, and $a_t$ is the expert action target. The future target chunk is $A_t=(a_t,\ldots,a_{t+K-1})$, with invalid entries masked at trajectory boundaries.
The initial layout is shared, while the target and subsequent motion depend on $\ell$.
We retain a generated family only when all instruction variants succeed.
Table~\ref{tab:components} specifies each task's paired additions.
Pairing provides different decisions under the same initial visual context.

\subsection{Physics-consistent visual expansion}
Re-rendering uses the benchmark's ManiSkill3/SAPIEN simulator~\citep{tao2024maniskill3,xiang2020sapien}, following simulation-based visual randomization~\citep{tobin2017domain}.
For a saved physical trajectory $\tau$ (omitting its scene and instruction subscripts), the renderer $\mathcal R$ restores each state $x_t$ and produces base- and hand-camera images under visual parameters $\eta$, which specify camera and lighting settings:
\begin{equation}
 I_t^{\eta}=\mathcal R(x_t;\eta),\qquad
 (s_t^{\eta},A_t^{\eta})=(s_t,A_t).
 \label{eq:render}
\end{equation}
Only observation formation changes; proprioception and future action targets remain tied to the saved physical states.
Simulator rendering preserves viewpoint-dependent geometry and action-label alignment.
Camera and illumination profiles apply to all five tasks; T5 also uses combined camera and lighting perturbations (Table~\ref{tab:components}).
Camera L1/L2/L3 use translation bounds of $0.03/0.06/0.12$\,m and rotation bounds of $2/6/12^\circ$; lighting half-widths are $0.10/0.25/0.40$ around ambient RGB 0.5.
T5 samples clean/camera/light/combined variants with probabilities $0.45/0.30/0.20/0.05$; combined rendering covers 120 base trajectories.

Expansion can otherwise overweight trajectories with many rendered copies.
For T5, the data loader samples an augmentation category $c$ (clean, camera, lighting, or combined), instruction $\ell$, compatible base-trajectory index $b$, rendered variant $v$, and valid training-window start $t$ hierarchically. The joint probability of selecting this training window is
\begin{equation}
 p(c,\ell,b,v,t)=
 \frac{w_c}{|\mathcal L_c|\,|\mathcal B_{c,\ell}|\,|\mathcal V_{c,b}|\,|\mathcal T_{b,v}|}.
 \label{eq:sampling}
\end{equation}
\begin{wraptable}{r}{0.60\textwidth}
\centering
\caption{Task coverage and execution settings.}
\label{tab:components}
\small
\setlength{\tabcolsep}{3pt}
\begin{tabularx}{\linewidth}{@{}lYrrrr@{}}
\toprule
Task & Paired instructions & Records & $H$ & $D$ & $\theta_g$ (mm)\\
\midrule
T1 & cap / body & 6,400 & 10 & 5 & 35\\
T2 & forward / left / right & 3,300 & 25 & 2 & 15\\
T3 & red / blue switch & 2,080 & 25 & 5 & ---\\
T4 & red / blue / green cube & 2,940 & 25 & 5 & 35\\
T5 & C / o / R / L & 7,840 & 25 & 30 & 35\\
\bottomrule
\end{tabularx}
\par\smallskip
{\footnotesize\raggedright All tasks use pairing and camera/light L1--L3; only T5 additionally uses combined camera and lighting perturbations. Records include rendered variants. $H$: execution horizon; $D$: denoising steps; $\theta_g$: closure threshold (---: continuous).\par}
\end{wraptable}

Here $w_c=p(c)$ is the prescribed category probability. The sets $\mathcal L_c$, $\mathcal B_{c,\ell}$, $\mathcal V_{c,b}$, and $\mathcal T_{b,v}$ contain, respectively, eligible instructions in category $c$, base trajectories for $(c,\ell)$, rendered variants of trajectory $b$ in category $c$, and valid window starts in variant $(b,v)$; $|\cdot|$ denotes set cardinality. Each choice after $c$ is uniform within its conditional eligible set. Thus, $p$ is the training-data sampling distribution, rather than the learned policy's action probability. Sampling base trajectories before rendered variants prevents extra visual copies from increasing a trajectory's sampling weight within a category and instruction.
The final T5 pool contains 2,800 distinct successful trajectories and 7,840 records after rendering, including 2,000 new motions from 500 four-instruction scenes.

\subsection{Supervised VLA adaptation}
We adapt Xiaomi-Robotics-1~\citep{xiaomi} with the task instruction, base and hand-eye RGB images, and joint/gripper state.
The native state packs seven arm joint positions and the mean of two finger positions.
The policy predicts $K=30$ end-effector actions, with unused action slots masked.
We retain Xiaomi-Robotics-1's architecture and released training implementation~\citep{xiaomi}, including masks, weighting, and auxiliary terms; action generation uses flow matching~\citep{lipman2023flow}.
Vision, language, action, and projection modules are updated with initialization-matched normalization and observation--action alignment.
The final T5 continuation uses 12,000 BF16 updates (batch 4), a cosine learning-rate schedule from $3\times10^{-6}$ to $5\times10^{-7}$, and 300 warmup updates.

\subsection{Task-aware closed-loop execution}
Local end-effector offsets are transformed into world-frame targets and converted into seven joint increments by inverse kinematics.
Finger targets are clipped to $[-0.01,0.04]$\,m; values below the task threshold $\theta_g$ map to full closure ($-0.01$\,m).
T3 retains continuous gripper targets.
Following action-chunking policies~\citep{zhao2023learning}, the robot executes the first $H\leq K$ actions before obtaining fresh observations, as in receding-horizon control~\citep{chi2023diffusion}.
We select $H$, denoising count $D$, and closure threshold $\theta_g$ through task-specific evaluation rather than imposing one configuration on all contacts (Table~\ref{tab:components}).

\section{Experiments}
\subsection{Setup and official results}
\begin{figure}[!ht]
\centering
\includegraphics[width=0.88\linewidth]{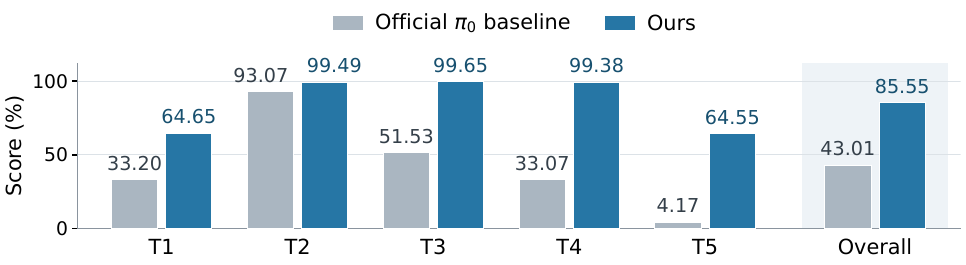}
\vspace{-4pt}
\caption{Comparison with the official $\pi_0$ baseline on the final leaderboard~\citep{leaderboard}.}
\label{fig:baseline}
\end{figure}

For each task, we initialize the policy from a pretrained Xiaomi-Robotics-1 checkpoint and fine-tune it on task-specific demonstrations.
Evaluation follows MetaFine's Perception ($P$), Understanding ($U$), and Controlled Behavior ($B$) protocol~\citep{xu2026beyond,benchmark}.
Here $P$ denotes perception AUSC (area under the success curve). Task scores average $P,U,B$, and Overall averages the five task scores.
Table~\ref{tab:results} reports the organizer's final results~\citep{leaderboard}.
DEED achieves 85.55\% Overall, versus 43.01\% for the published $\pi_0$ baseline (+42.54 percentage points; Figure~\ref{fig:baseline})~\citep{leaderboard}.

\begin{table}[!ht]
\begin{minipage}[t]{0.48\linewidth}
\centering
\caption{Official final scores (\%).}
\label{tab:results}
\small
\begin{tabular}{lrrrr}
\toprule
Task & $P$ & $U$ & $B$ & Mean\\
\midrule
T1 & 68.96 & 62.50 & 62.50 & 64.65\\
T2 & 98.47 & 100.00 & 100.00 & 99.49\\
T3 & 98.96 & 100.00 & 100.00 & 99.65\\
T4 & 98.15 & 100.00 & 100.00 & 99.38\\
T5 & 61.15 & 66.25 & 66.25 & 64.55\\
\midrule
Average & 85.14 & 85.75 & 85.75 & \textbf{85.55}\\
\bottomrule
\end{tabular}
\end{minipage}\hfill
\begin{minipage}[t]{0.49\linewidth}
\centering
\caption{Local execution ablations (\%; bold: deployed settings).}
\label{tab:ablation}
\small
\begin{tabular}{llrrr}
\toprule
Task & Setting & $P$ & $U$ & SR\\
\midrule
T1 & $H=25$ & 84.03 & 69.17 & 80.93\\
 & $\boldsymbol{H=10}$ & \textbf{93.75} & \textbf{78.33} & \textbf{90.19}\\
 & $H=5$ & 85.56 & 65.83 & 81.30\\
\midrule
T3 & $D=3$ & 98.13 & 97.50 & 98.06\\
 & $\boldsymbol{D=5}$ & \textbf{98.13} & \textbf{97.50} & \textbf{98.06}\\
 & $D=10$ & 97.29 & 95.00 & 96.67\\
\bottomrule
\end{tabular}
\end{minipage}
\end{table}

\vspace{-6pt}
\subsection{Fixed-policy component ablations}
Table~\ref{tab:ablation} uses the latest completed execution sweeps for the retained policies, with weights and normalization fixed within each comparison and flow-noise seed 0.
The T1 sweep comprises three batches of 20 scenes, totaling 60 scenes and 540 episodes per horizon; the T3 sweep uses 40 new scenes and 360 episodes per denoising setting.
SR pools two clean instruction episodes and seven perception episodes per scene with equal episode weights.
Both sweeps preserve the public evaluator's environment reuse: T1 runs perception followed by understanding within each batch, whereas T3 evaluates the two modes in separate processes.
Scene seeds are shared across settings, but subsequent physical initial states can differ because of environment history; these are aggregate comparisons rather than fully initial-state-matched counterfactuals.
Changing $H$ also changes the chunk-indexed flow-noise schedule.

In the T1 sweep, $H=10$ scores higher than both $H=25$ and $H=5$; decreasing the execution window further is therefore not consistently beneficial.
On T3, $D=3$ and the deployed $D=5$ tie, while $D=10$ is lower by 0.83 points in $P$ and 2.50 points in $U$. The expanded sweep did not include $D=1$. Because the local sweep showed no degradation, $D=5$ was submitted for organizer evaluation and retained for deployment.
A 20-scene gripper study on an earlier partially adapted T1 policy yields $P=27.92\%$ and $U=14/40$ (35.00\%) with continuous targets, versus $P=57.92\%$ and $U=22/40$ (55.00\%) with full closure.
Across eight expert-screened T5 scenes, correct instructions succeed in 21/32 trials (65.63\%) versus 0/32 (0\%) under cyclic letter permutation, with all initial-state and image pairs matched.

\vspace{-8pt}
\section{Conclusion}
%Our local sweeps favor task-specific execution settings: neither shorter execution windows nor larger denoising budgets consistently improve performance. The T5 diagnostic further shows that correct language input matters even when physical initial states and images are matched. Future studies should isolate the contributions of demonstration pairing, visual expansion, and hierarchical sampling, and assess their stability across broader scene distributions.

We present DEED, a framework that combines demonstration expansion with task-aware closed-loop execution for fine-grained manipulation. DEED generates instruction and visual variations while preserving state-action consistency, and translates predicted action chunks into robot control through inverse kinematics, task-specific gripper mappings, and receding-horizon replanning. It achieves an official Overall score of 85.55\% across the five MetaFine tasks, while fixed-policy evaluations highlight the importance of execution configuration.

\clearpage
\bibliography{references}

@article{xu2026beyond,
  title={Beyond Binary Success: A Diagnostic Meta-Evaluation Framework for Fine-Grained Manipulation},
  author={Xu, He-Yang and Zhang, Pengyuan and Ge, Zongyuan and Hao, Xiaoshuai and Belongie, Serge and Geng, Xin and Peng, Yuxin and Wei, Xiu-Shen},
  journal={arXiv preprint arXiv:2605.19986},
  year={2026},
  url={https://arxiv.org/abs/2605.19986}
}

@misc{benchmark,
  author = {{Hiangx Robotics}},
  title = {{MetaFine-Challenge\_CoRL}: Participant Guide and Evaluation Protocol},
  year = {2026},
  howpublished = {\url{https://github.com/Hiangx-robotics/MetaFine-Challenge_CoRL}}
}

@misc{dataset,
  author = {{Hiangx Robotics}},
  title = {{MetaFine\_CoRL26} Demonstration Dataset},
  year = {2026},
  howpublished = {\url{https://huggingface.co/datasets/hiangx/MetaFine_CoRL26}}
}

@article{xiaomi,
  title={{Xiaomi-Robotics-1: Scaling Vision-Language-Action Models with over 100K Hours of Real-World Trajectories}},
  author={{Xiaomi Robotics Team} and Jun Guo and Piaopiao Jin and Jason Li and Peiyan Li and Yingyan Li and Futeng Liu and Wanli Peng and Optimus Qin and Yifei Su and Nan Sun and Qiao Sun and Runze Suo and Heyun Wang and Yunhong Wang and Rujie Wu and Caoyu Xia and Lina Zhang and Jack Zhao and Guoliang Chen and Wenlong Chen and Xinze He and Bin Li and Qing Li and Zhuorong Li and Heng Qu and Wenxuan Song and Diyun Xiang and Yifan Xie and Peiran Xu and Hangjun Ye and Wen Ye and Han Zhao and Quanyun Zhou},
  journal={arXiv preprint arXiv:2607.15330},
  year={2026},
  url={https://arxiv.org/abs/2607.15330},
  note={Code: \url{https://github.com/XiaomiRobotics/Xiaomi-Robotics-1}}
}

@misc{leaderboard,
  author = {{RoboFineMani Organizers}},
  title = {{RoboFineMani 2026 Official Leaderboard}},
  year = {2026},
  howpublished = {\url{https://robofinemani2026.github.io/leaderboard.html}},
  note = {Updated 24 October 2026; accessed 28 October 2026}
}

@inproceedings{shridhar2022cliport,
 title={{CLIPort}: What and Where Pathways for Robotic Manipulation},
 author={Shridhar, Mohit and Manuelli, Lucas and Fox, Dieter},
 booktitle={Proceedings of the 5th Conference on Robot Learning},
 series={Proceedings of Machine Learning Research}, volume={164}, pages={894--906}, year={2022},
 url={https://proceedings.mlr.press/v164/shridhar22a.html}
}

@article{tobin2017domain,
 title={Domain Randomization for Transferring Deep Neural Networks from Simulation to the Real World},
 author={Tobin, Josh and Fong, Rachel and Ray, Alex and Schneider, Jonas and Zaremba, Wojciech and Abbeel, Pieter},
 journal={arXiv preprint arXiv:1703.06907}, year={2017},
  url={https://arxiv.org/abs/1703.06907}
}

@article{zhao2023learning,
 title={Learning Fine-Grained Bimanual Manipulation with Low-Cost Hardware},
 author={Zhao, Tony Z. and Kumar, Vikash and Levine, Sergey and Finn, Chelsea},
 journal={arXiv preprint arXiv:2304.13705}, year={2023},
  url={https://arxiv.org/abs/2304.13705}
}

@inproceedings{chi2023diffusion,
 title={Diffusion Policy: Visuomotor Policy Learning via Action Diffusion},
 author={Chi, Cheng and Feng, Siyuan and Du, Yilun and Xu, Zhenjia and Cousineau, Eric and Burchfiel, Benjamin CM and Song, Shuran},
 booktitle={Proceedings of Robotics: Science and Systems}, year={2023},
 doi={10.15607/RSS.2023.XIX.026},
  url={https://doi.org/10.15607/RSS.2023.XIX.026}
}

@inproceedings{lipman2023flow,
 title={Flow Matching for Generative Modeling},
 author={Lipman, Yaron and Chen, Ricky T. Q. and Ben-Hamu, Heli and Nickel, Maximilian and Le, Matt},
 booktitle={International Conference on Learning Representations}, year={2023},
 url={https://arxiv.org/abs/2210.02747}
}

@article{brohan2023rt2,
 title={{RT-2: Vision-Language-Action Models Transfer Web Knowledge to Robotic Control}},
 author={Anthony Brohan and Noah Brown and Justice Carbajal and Yevgen Chebotar and Xi Chen and Krzysztof Choromanski and Tianli Ding and Danny Driess and Avinava Dubey and Chelsea Finn and Pete Florence and Chuyuan Fu and Montse Gonzalez Arenas and Keerthana Gopalakrishnan and Kehang Han and Karol Hausman and Alexander Herzog and Jasmine Hsu and Brian Ichter and Alex Irpan and Nikhil Joshi and Ryan Julian and Dmitry Kalashnikov and Yuheng Kuang and Isabel Leal and Lisa Lee and Tsang-Wei Edward Lee and Sergey Levine and Yao Lu and Henryk Michalewski and Igor Mordatch and Karl Pertsch and Kanishka Rao and Krista Reymann and Michael Ryoo and Grecia Salazar and Pannag Sanketi and Pierre Sermanet and Jaspiar Singh and Anikait Singh and Radu Soricut and Huong Tran and Vincent Vanhoucke and Quan Vuong and Ayzaan Wahid and Stefan Welker and Paul Wohlhart and Jialin Wu and Fei Xia and Ted Xiao and Peng Xu and Sichun Xu and Tianhe Yu and Brianna Zitkovich},
 journal={arXiv preprint arXiv:2307.15818},
 year={2023},
 url={https://arxiv.org/abs/2307.15818}
}

@article{kim2024openvla,
 title={{OpenVLA: An Open-Source Vision-Language-Action Model}},
 author={Moo Jin Kim and Karl Pertsch and Siddharth Karamcheti and Ted Xiao and Ashwin Balakrishna and Suraj Nair and Rafael Rafailov and Ethan Foster and Grace Lam and Pannag Sanketi and Quan Vuong and Thomas Kollar and Benjamin Burchfiel and Russ Tedrake and Dorsa Sadigh and Sergey Levine and Percy Liang and Chelsea Finn},
 journal={arXiv preprint arXiv:2406.09246},
 year={2024},
 url={https://arxiv.org/abs/2406.09246}
}

@article{octo2024,
 title={{Octo: An Open-Source Generalist Robot Policy}},
 author={{Octo Model Team} and Dibya Ghosh and Homer Walke and Karl Pertsch and Kevin Black and Oier Mees and Sudeep Dasari and Joey Hejna and Tobias Kreiman and Charles Xu and Jianlan Luo and You Liang Tan and Lawrence Yunliang Chen and Pannag Sanketi and Quan Vuong and Ted Xiao and Dorsa Sadigh and Chelsea Finn and Sergey Levine},
 journal={arXiv preprint arXiv:2405.12213},
 year={2024},
 url={https://arxiv.org/abs/2405.12213}
}

@article{black2024pi0,
 title={{$\pi_0$: A Vision-Language-Action Flow Model for General Robot Control}},
 author={Kevin Black and Noah Brown and Danny Driess and Adnan Esmail and Michael Equi and Chelsea Finn and Niccolo Fusai and Lachy Groom and Karol Hausman and Brian Ichter and Szymon Jakubczak and Tim Jones and Liyiming Ke and Sergey Levine and Adrian Li-Bell and Mohith Mothukuri and Suraj Nair and Karl Pertsch and Lucy Xiaoyang Shi and James Tanner and Quan Vuong and Anna Walling and Haohuan Wang and Ury Zhilinsky},
 journal={arXiv preprint arXiv:2410.24164},
 year={2024},
 url={https://arxiv.org/abs/2410.24164}
}

@article{openx2023,
 title={{Open X-Embodiment: Robotic Learning Datasets and RT-X Models}},
 author={{Open X-Embodiment Collaboration}},
 journal={arXiv preprint arXiv:2310.08864},
 year={2023},
 url={https://arxiv.org/abs/2310.08864}
}

@article{mandlekar2023mimicgen,
 title={{MimicGen: A Data Generation System for Scalable Robot Learning using Human Demonstrations}},
 author={Ajay Mandlekar and Soroush Nasiriany and Bowen Wen and Iretiayo Akinola and Yashraj Narang and Linxi Fan and Yuke Zhu and Dieter Fox},
 journal={arXiv preprint arXiv:2310.17596},
 year={2023},
 url={https://arxiv.org/abs/2310.17596}
}

@article{nasiriany2024robocasa,
 title={{RoboCasa: Large-Scale Simulation of Everyday Tasks for Generalist Robots}},
 author={Soroush Nasiriany and Abhiram Maddukuri and Lance Zhang and Adeet Parikh and Aaron Lo and Abhishek Joshi and Ajay Mandlekar and Yuke Zhu},
 journal={arXiv preprint arXiv:2406.02523},
 year={2024},
 url={https://arxiv.org/abs/2406.02523}
}

@article{tao2024maniskill3,
 title={{ManiSkill3: GPU Parallelized Robotics Simulation and Rendering for Generalizable Embodied AI}},
 author={Stone Tao and Fanbo Xiang and Arth Shukla and Yuzhe Qin and Xander Hinrichsen and Xiaodi Yuan and Chen Bao and Xinsong Lin and Yulin Liu and Tse-kai Chan and Yuan Gao and Xuanlin Li and Tongzhou Mu and Nan Xiao and Arnav Gurha and Viswesh Nagaswamy Rajesh and Yong Woo Choi and Yen-Ru Chen and Zhiao Huang and Roberto Calandra and Rui Chen and Shan Luo and Hao Su},
 journal={arXiv preprint arXiv:2410.00425},
 year={2024},
 url={https://arxiv.org/abs/2410.00425}
}

@article{xiang2020sapien,
 title={{SAPIEN: A SimulAted Part-based Interactive ENvironment}},
 author={Fanbo Xiang and Yuzhe Qin and Kaichun Mo and Yikuan Xia and Hao Zhu and Fangchen Liu and Minghua Liu and Hanxiao Jiang and Yifu Yuan and He Wang and Li Yi and Angel X. Chang and Leonidas J. Guibas and Hao Su},
 journal={arXiv preprint arXiv:2003.08515},
 year={2020},
 url={https://arxiv.org/abs/2003.08515}
}

@article{mees2021calvin,
 title={{CALVIN: A Benchmark for Language-Conditioned Policy Learning for Long-Horizon Robot Manipulation Tasks}},
 author={Oier Mees and Lukas Hermann and Erick Rosete-Beas and Wolfram Burgard},
 journal={arXiv preprint arXiv:2112.03227},
 year={2021},
 url={https://arxiv.org/abs/2112.03227}
}
\end{document}